\documentclass[10pt,twocolumn,letterpaper]{article}

\usepackage[pagenumbers]{cvpr} % To force page numbers, e.g. for an arXiv version

\usepackage{times}
\usepackage{epsfig}
\usepackage{graphicx}
\usepackage{amsmath}
\usepackage{amssymb}

\usepackage{enumitem}
\usepackage{booktabs}
\usepackage{graphicx}

\makeatletter
\DeclareRobustCommand\onedot{\futurelet\@let@token\@onedot}
\def\@onedot{\ifx\@let@token.\else.\null\fi\xspace}

\def\eg{\emph{e.g}\onedot} 
\def\ie{\emph{i.e}\onedot} 
\def\cf{\emph{cf}\onedot}

\def\etal{\emph{et al}\onedot}

\newcommand{\figref}[1]{Fig\onedot~\ref{#1}}
\newcommand{\equref}[1]{Eq\onedot~\eqref{#1}}
\newcommand{\secref}[1]{Sec\onedot~\ref{#1}}
\newcommand{\tabref}[1]{Tab\onedot~\ref{#1}}
\renewcommand{\paragraph}[1]{\vspace{1mm}\noindent\textbf{#1}}
\usepackage{pifont}

\newlength\savewidth\newcommand\shline{\noalign{\global\savewidth\arrayrulewidth
  \global\arrayrulewidth 1pt}\hline\noalign{\global\arrayrulewidth\savewidth}}
\newcommand{\tablestyle}[2]{\setlength{\tabcolsep}{#1}\renewcommand{\arraystretch}{#2}\centering\footnotesize}

\usepackage{xcolor}
\usepackage{tabularx}
\usepackage{multirow}
\makeatletter
\newcommand{\thickhline}{%
\noalign{\ifnum0=`}\fi\hrule \@height 1.5pt %
\futurelet\reserved@a\@xhline}
\makeatother

\usepackage{bm}
\newcommand{\hatd}{\hat{d}}
\newcommand{\haty}{\hat{y}}
\newcommand{\hatp}{\hat{p}}
\newcommand{\hatm}{\hat{m}}
\newcommand{\hatc}{\hat{c}}

\newcommand{\hz}{\hat{z}}

\newcommand{\hy}{\hat{y}}

\newcommand{\hp}{\hat{p}}

\renewcommand{\Sigma}{\mathfrak{S}}

\def\1{\bm{1}}

\DeclareMathAlphabet{\mathsfit}{\encodingdefault}{\sfdefault}{m}{sl}
\SetMathAlphabet{\mathsfit}{bold}{\encodingdefault}{\sfdefault}{bx}{n}

\def\sC{{\mathbb{C}}}

\DeclareMathOperator*{\argmax}{arg\,max}

\usepackage[pagebackref,breaklinks,colorlinks]{hyperref}
\usepackage[capitalize]{cleveref}
\crefname{section}{Sec.}{Secs.}
\Crefname{section}{Section}{Sections}
\Crefname{table}{Table}{Tables}
\crefname{table}{Tab.}{Tabs.}

\def\cvprPaperID{3981} % *** Enter the CVPR Paper ID here
\def\confName{CVPR}
\def\confYear{2022}

\begin{document}
\title{TubeFormer-DeepLab: Video Mask Transformer}

\author{
Dahun Kim\textsuperscript{1,3}~~~~Jun Xie\textsuperscript{3}~~~~Huiyu Wang\textsuperscript{2}~~~Siyuan Qiao\textsuperscript{3}~~~~Qihang Yu\textsuperscript{2}~~~~~~Hong-Seok Kim\textsuperscript{3}\\
Hartwig Adam\textsuperscript{3}~~~~~~In So Kweon\textsuperscript{1}~~~~~~Liang-Chieh Chen\textsuperscript{3}\\\textsuperscript{1}KAIST~~~~~\textsuperscript{2}Johns Hopkins University~~~~\textsuperscript{3}Google Research}

\maketitle

%%%%%%%%% ABSTRACT
\vspace{-2mm}
\begin{abstract}
\vspace{-2mm}
We present TubeFormer-DeepLab, the first attempt to tackle multiple core video segmentation tasks in a unified manner.
Different video segmentation tasks (\eg, video semantic/instance/panoptic segmentation) are usually considered as distinct problems.
State-of-the-art models adopted in the separate communities have diverged, and radically different approaches dominate in each task.
By contrast, we make a crucial observation that video segmentation tasks could be generally formulated as the problem of assigning different predicted labels to video tubes (where a tube is obtained by linking segmentation masks along the time axis) and the labels may encode different values depending on the target task.
The observation motivates us to develop TubeFormer-DeepLab, a simple and effective video mask transformer model that is widely applicable to multiple video segmentation tasks.
TubeFormer-DeepLab directly predicts video tubes with task-specific labels (either pure semantic categories, or both semantic categories and instance identities), which not only significantly simplifies video segmentation models, but also advances state-of-the-art results on multiple video segmentation benchmarks.
\vspace{-2mm}
\end{abstract}

%%%%%%%%% BODY TEXT
\section{Introduction}
\label{sec:intro}

\begin{figure}[t]
\vspace{-0.5mm}
\centering
\includegraphics[width=0.75\linewidth]{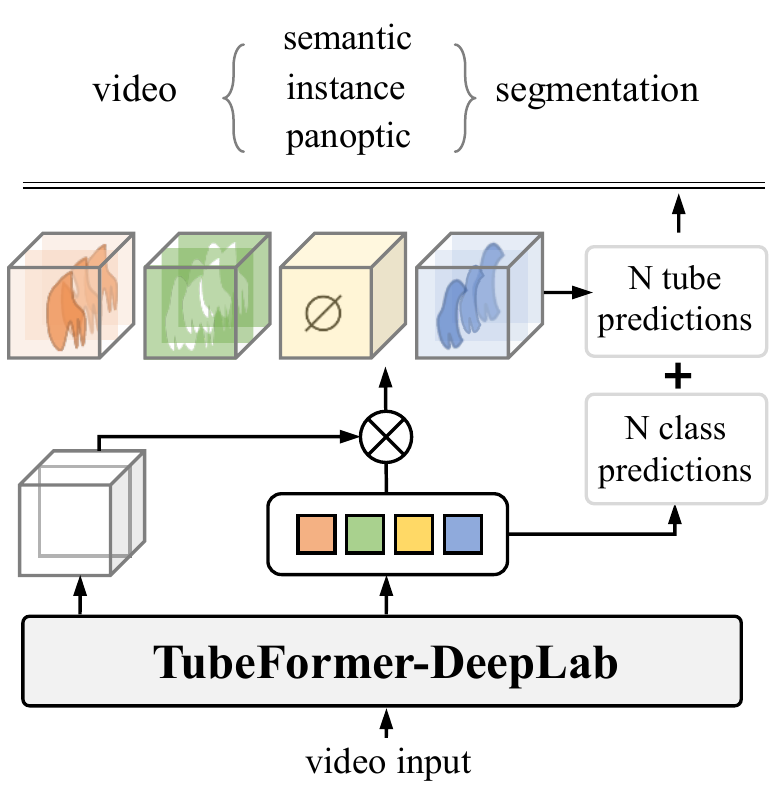}
\vspace{-2.2mm}
\caption{\small Video segmentation tasks can be formulated as partitioning video frames (\eg, a clip) into tubes (\ie, segmentation masks linked along time) with different labels. TubeFormer-DeepLab directly predicts class-labeled tubes, providing a simple and general solution to Video Semantic Segmentation (VSS), Video Instance Segmentation (VIS), and Video Panoptic Segmentation (VPS).
}
\label{fig:teaser}
\vspace{-1.0mm}
\end{figure}

\begin{figure}[t]
\centering
\includegraphics[width=0.9\linewidth]{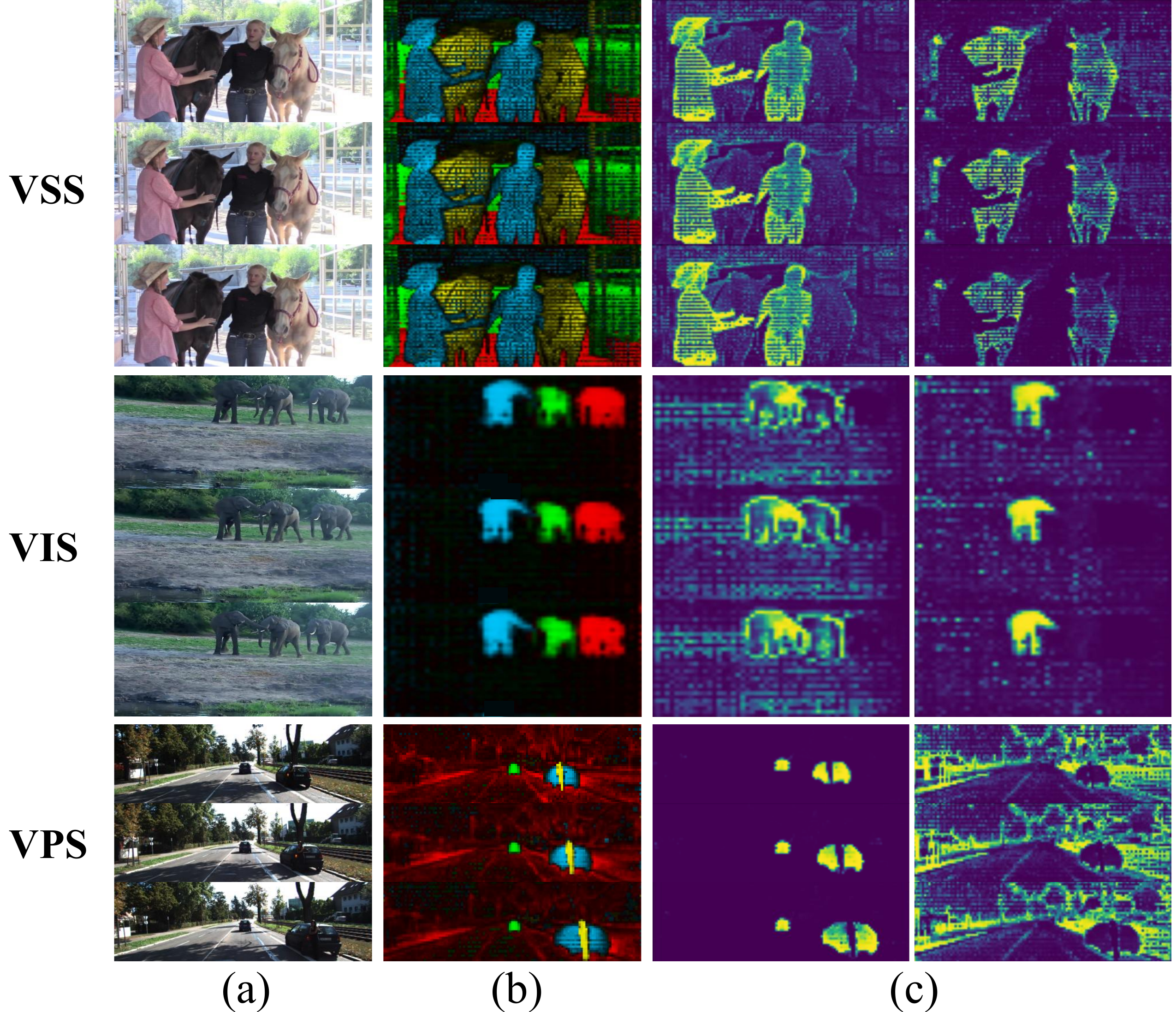}
\vspace{-2.5mm}
\caption{\small Our proposed hierarchical dual-path transformer performs attention on three consecutive input frames (a) for VSS, VIS, and VPS tasks. While the global memory learns the spatio-temporally clustered attention for individual tube regions (b), our latent memory learns task-specific attention (c).}
\label{fig:teaser_attention}
\vspace{-10mm}
\end{figure}

We observe that video segmentation tasks could be formulated as {\it partitioning video frames into tubes with different predicted labels}, where a tube contains segmentation masks linked along the time axis.
Based on the target task, the predicted labels may encode only semantic categories (\eg, Video Semantic Segmentation (VSS)~\cite{BrostowSFC:ECCV08,miao2021vspw}), or both semantic categories and instance identities (\eg, Video Instance Segmentation (VIS)~\cite{voigtlaender2019mots,Yang19ICCV} for only foreground `things', or Video Panoptic Segmentation (VPS)~\cite{kim2020video,Weber2021NEURIPSDATA} for both foreground `things' and background `stuff') (\figref{fig:teaser}).

However, the underlying similarity of several video segmentation tasks (\ie, assigning tubes with predicted labels) has been long overlooked, and thus models developed for video semantic, instance, and panoptic segmentation have fundamentally diverged. 
For example, some VSS methods~\cite{zhu2017deep,gadde2017semantic} warp features between video frames, while the modern VIS model~\cite{bertasius2020classifying} predicts hundreds of frame-level instance masks~\cite{he2017mask} and then propagates them to other neighboring frames. To make matters more complicated, state-of-the-art VPS methods~\cite{qiao2021detectors,woo2021learning} adopt separate prediction branches, specific to semantic segmentation, instance segmentation, and object tracking, respectively.

In this work, instead of exacerbating the bifurcation between video segmentation models, we take a step back and rethink the following question: {\it Can we exploit the similar nature between video segmentation tasks, and develop a single model that is both effective and generally applicable?} To answer this, we propose {\bf TubeFormer-DeepLab} that builds upon mask transformers~\cite{wang2021max} for video segmentation by directly predicting class-labeled tubes, where the labels encode different values depending on the target task.

Specifically, similar to other Transformer architectures~\cite{vaswani2017attention,carion2020end}, TubeFormer-DeepLab extends the mask transformer~\cite{wang2021max} to generate a set of pairs, each containing a class prediction and a tube embedding vector. The tube embedding vector, multiplied by the video pixel embedding features obtained by a convolutional network~\cite{lecun1998gradient}, yields the tube prediction.
As a result, TubeFormer-DeepLab presents the first attempt to tackle multiple core video segmentation tasks in a general framework without the need to adapt the system for any task-specific design.

Na\"ively applying the image-level mask transformer~\cite{wang2021max} to the video domain does not yield a satisfactory result, mainly due to the difficulty of learning attentions for {\it video-clip} (\ie, multi-frames) features with large spatial resolutions. To alleviate the issue, we introduce the {\bf latent} dual-path transformer block that is in charge of passing messages between {\it video-frame} (\ie, single-frame) features and a {\bf latent} memory, followed by the {\bf global} dual-path transformer block that learns the attentions between {\it video-clip} features and a {\bf global} memory. This hierarchical dual-path transformer framework facilitates the attention learning and significantly improves the video segmentation results.
Interestingly, as shown in~\figref{fig:teaser_attention}, our latent memory learns task-specific attention, while the global memory learns the spatio-temporally clustered attention for individual tube regions.
Additionally, we split the global memory into two sets, thing-specific and stuff-specific global memory, with the motivation to exploit the different nature of `thing' (countable instances) and `stuff' (amorphous regions).

During inference, practically we could only fit a video clip (\ie, a short video sequence) for video segmentation. The whole video sequence segmentation result is thus obtained by applying the video stitching~\cite{qiao2021vip} to merge clip segmentation results. To enforce the consistency between video clips, we additionally propose a Temporal Consistency loss that encourages the model to learn consistent predictions in the overlapping frames between clips.

Finally, we propose a simple and effective data augmentation policy by extending the image-level thing-specific copy-paste~\cite{fang2019instaboost,ghiasi2021simple}. Our  method, named clip-paste (clip-level copy-paste), randomly pastes either `thing' or `stuff' (or both) regions from a video clip to the target video clip.

To demonstrate the effectiveness of our proposed TubeFormer-DeepLab, we conduct experiments on multiple core video segmentation datasets, including KITTI-STEP (VPS)~\cite{Weber2021NEURIPSDATA}, VSPW (VSS)~\cite{miao2021vspw}, YouTube-VIS (VIS)~\cite{Yang19ICCV},  SemKITTI-DVPS (depth-aware VPS)~\cite{qiao2021vip}, and recent VIPSeg~\cite{miao2022large} (VPS). Our {\it single} model not only significantly simplifies video segmentation systems (\eg, the proposed model is end-to-end trained and does not require any task-specific design), but also advances state-of-the-art performance on several benchmarks.
In particular, TubeFormer-DeepLab outperforms {\it published works} Motion-DeepLab~\cite{Weber2021NEURIPSDATA} by {\bf +13.1} STQ on KITTI-STEP {\it test} set, TCB~\cite{miao2021vspw} by {\bf +21} mIoU on VSPW {\it test} set, IFC~\cite{hwang2021video} by {\bf +2.9} track-mAP on YouTube-VIS-2019 {\it val} set, ViP-DeepLab~\cite{qiao2021vip} by {\bf +3.6} DSTQ on SemKITTI-DVPS {\it test} set,
Clip-PanoFCN~\cite{miao2022large} by {\bf +13.6} STQ and {\bf +3.9} VPQ on VIPSeg {\it test} set.
Our experimental results validate TubeFormer-DeepLab's general efficacy for video segmentation tasks. 

\section{Related Works}
\label{sec:related}

{\noindent \bf Video Semantic Segmentation (VSS).\quad} Extending image semantic segmentation~\cite{he2004multiscale,everingham2010pascal,long2014fully,deeplabv12015,cordts2016cityscapes,zhao2017pyramid,deeplabv3plus2018,xiao2018unified,zhu2019asymmetric,huang2019ccnet,yuan2020object} to the video domain requires predicting all pixels in a video with different semantic classes~\cite{BrostowSFC:ECCV08,miao2021vspw}. Prior methods~\cite{zhu2017deep,gadde2017semantic,nilsson2018semantic,li2018low,zhu2019improving,jain2019accel} exploit the temporal information via a warping module~\cite{horn1981determining,jaderberg2015spatial,dai2017deformable}. Recently, Mao \etal~\cite{miao2021vspw} introduced a large-scale VSS benchmark, called VSPW (Video Scene Parsing in the Wild), along with a solid baseline that effectively aggregates video context information by extending ~\cite{yuan2020object} and ~\cite{zhao2017pyramid} to the temporal dimension.

{\noindent\bf Video Instance Segmentation (VIS). \quad} Combining multi-object tracking~\cite{Breitenstein2009ICCV,geiger2013vision,bergmann2019tracking,peng2020chained,dendorfer2020ijcv} and instance segmentation~\cite{hariharan2014simultaneous,he2017mask,liu2018path,chen2019hybrid,tian2020conditional,qiao2021detectors}, video instance segmentation~\cite{voigtlaender2019mots,Yang19ICCV} aims to track instance masks across video frames. Most state-of-the-art VIS  methods~\cite{Yang19ICCV,bertasius2020classifying,cao2020sipmask,liu2021sg,fu2020compfeat,wang2021end,lin2021video,hwang2021video} are detection-based approaches, allowing overlapping mask predictions (\eg, based on Mask R-CNN~\cite{he2017mask}, FCOS~\cite{tian2019fcos}, or DETR~\cite{carion2020end,zhu2020deformable}).
Our work is similar to the concurrent work IFC~\cite{hwang2021video}, which uses memory features for video instance segmentation. However, our work does not exploit memory features for inter-frame communication, and thus does not require extra modules to perform such a task. Instead, the latent memory features are deployed in the proposed latent dual-path transformer block to facilitate \textit{per-frame} segmentation. Finally, LatentGNN~\cite{zhang2019latentgnn} also explored the latent features in the graphical neural networks~\cite{scarselli2008graph}.

{\noindent\bf Video Panoptic Segmentation (VPS).\quad} Recently, panoptic segmentation~\cite{kirillov2018panoptic,kirillov2019panoptic,xiong2019upsnet,yang2019deeperlab,li2018attention,cheng2019panopticworkshop,wang2020axial,wang2021max,li2021fully} has also been extended to the video domain. Video Panoptic Segmentation~\cite{kim2020video} attempts to unify video semantic and instance segmentation, requiring temporally consistent panoptic segmentation results. Different from VIS, VPS disallows overlapping instance masks and requires labeling each pixel, including both ‘thing’ and ‘stuff’ pixels. Current state-of-the-art approaches~\cite{woo2021learning,qiao2021vip,Weber2021NEURIPSDATA} adopted complicated pipelines due to the intricate nature of VPS. Specifically, VPSNet~\cite{kim2020video} contains multiple task-specific heads, including Mask R-CNN~\cite{he2017mask}, deformable convolutions~\cite{dai2017deformable}, and MaskTrack~\cite{Yang19ICCV} for instance segmentation, semantic segmentation, and tracking, respectively, while ViP-DeepLab~\cite{qiao2021vip} extends Panoptic-DeepLab~\cite{cheng2019panoptic} (which employs dual-ASPP~\cite{chen2017deeplabv3} and dual-decoder structures specific to semantic and instance segmentation, respectively) by adding another next-frame instance segmentation branch. On the other hand, our approach significantly simplifies the current pipeline by employing mask transformers~\cite{wang2021max} to directly predict clip-level mask segmentation results. Finally, our proposed model could also be easily extended to the recent task of Depth-aware Video Panoptic Segmentation (DVPS)~\cite{qiao2021vip}, which further requires per-pixel depth estimation on top of VPS results.
We note that current with our work, Video K-Net~\cite{li2022video}, extending K-Net~\cite{zhang2021k}, also develops a unified framework for video panoptic segmentation.

\section{Method}
\label{sec:methods}

In this section, we introduce the formulation of several video segmentation tasks, followed by a general formulation that inspires our TubeFormer-DeepLab.
We then present its model design, training and inference strategies.

\subsection{Video Segmentation Formulation}
\label{sec:model}

Let us denote with $v \in \mathbb{R}^{T\times H\times {W}\times 3}$ an input video clip containing ${T}$ video frames of spatial size ${H}\times {W}$ (${T}$ could be equal to the video sequence length if memory allows).
The video clip is annotated with a set of class-labeled tubes (a tube is defined as segmentation masks linked along the time axis): $\{y_i\}_{i=1}^K = \{(m_i, c_i)\}_{i=1}^K \,$, where the $K$ ground truth tubes $m_i \in {\{0,1\}}^{{T}\times {H}\times {W}}$ do not overlap with each other, and $c_i$ denotes the ground truth class label of tube $m_i$.
Below, we briefly introduce several tasks.

{\bf Video Semantic Segmentation (VSS)} is typically formulated as per-video pixel classification, where the pixel features for classification are enriched by warping~\cite{zhu2017deep} or aggregating~\cite{miao2021vspw} features from neighboring frames. Formally, the model predicts the probability distribution over a predefined set of categories $\sC = \{1, ..., D\}$ for every video pixel: $\{\hatp_{i}|\hatp_{i} \in \Delta^{D}\}_{i=1}^{{T}\times {H} \times {W}}$, where $\Delta^{D}$ is the $D$-dimensional probability simplex. The final segmentation output $\haty$ is then obtained by taking its argmax (\ie, $\haty_i=\argmax_c \hatp_i(c), \forall i \in \{1, 2, \dots, {T}\times {H} \times {W}\}$).

{\bf Video Instance Segmentation (VIS)} requires to segment and temporally link object instances in the video. For each detected foreground `thing' $i$ in the video, the model predicts a video tube (\ie, video-level instance mask track) $\hatm_i \in [0, 1]^{{T}\times {H}\times {W}}$ with a probability distribution $\hatp_i$ over $\sC$ defined for {\it only} thing classes. Depending on the target dataset or evaluation metric, the model may generate {\it overlapping} video tubes (\eg, Youtbue-VIS~\cite{Yang19ICCV} adopts track-mAP, allowing overlapping predicted tubes, while KITTI-MOTS~\cite{voigtlaender2019mots} adopts HOTA~\cite{Luiten20IJCV}, disallowing so).

{\bf Video Panoptic Segmentation (VPS)} requires temporally consistent semantic and instance segmentation results for both `thing' and `stuff' classes. Specifically, the model predicts a set of {\it non-overlapping} video tubes $\{\hy_i\}_{i=1}^N = \{(\hatm_i, \hp_i(c))\}_{i=1}^N$, where $\hatm_i \in {[0,1]}^{T\times H\times W}$ denotes the predicted tube, and $\hp_i(c)$ denotes the probability of assigning class $c$ to tube $\hatm_i$ belonging to a predefined category set $\sC$ that contains both `thing' and `stuff' classes.

{\bf Depth-aware Video Panoptic Segmentation (DVPS)} builds on top of VPS by additionally requiring a model to estimate the depth value of each pixel. Similar to VPS output, the prediction has the following format: $\{\haty_i\}_{i=1}^N = \{(\hatm_i, \hatp_i(c), \hatd_i)\}_{i=1}^N$, where $\hatd_i \in {[0,d_{max}]}^{{T}\times {H}\times {W}}$ denotes the estimated depth value and $d_{max}$ is the maximum depth value specified in the target dataset. Accordingly, the dataset contains ground truth depth.

\paragraph{General task formulation.}\quad
Despite the superficial differences between tasks, we discover the underlying similarity that video segmentation tasks could be generally formulated as the problem of assigning different predicted labels to video tubes and the labels may encode different values depending on the target task. For example, if only semantic categories are predicted, it becomes video semantic segmentation. Similarly, if both semantic categories and instance identities are required (\ie, one predicted tube for each category-identity pair), it then becomes either video instance segmentation (if only foreground `thing' classes are considered) or video panoptic segmentation. This motivates us to develop a general video segmentation model that directly predicts class-labeled tubes $\{\hy_i\}_{i=1}^N = \{(\hatm_i, \hp_i(c))\}_{i=1}^N$ (and optionally depth, if required).

\subsection{TubeFormer-DeepLab Architecture}

\begin{figure*}
\centering
\includegraphics[width=0.85\linewidth]{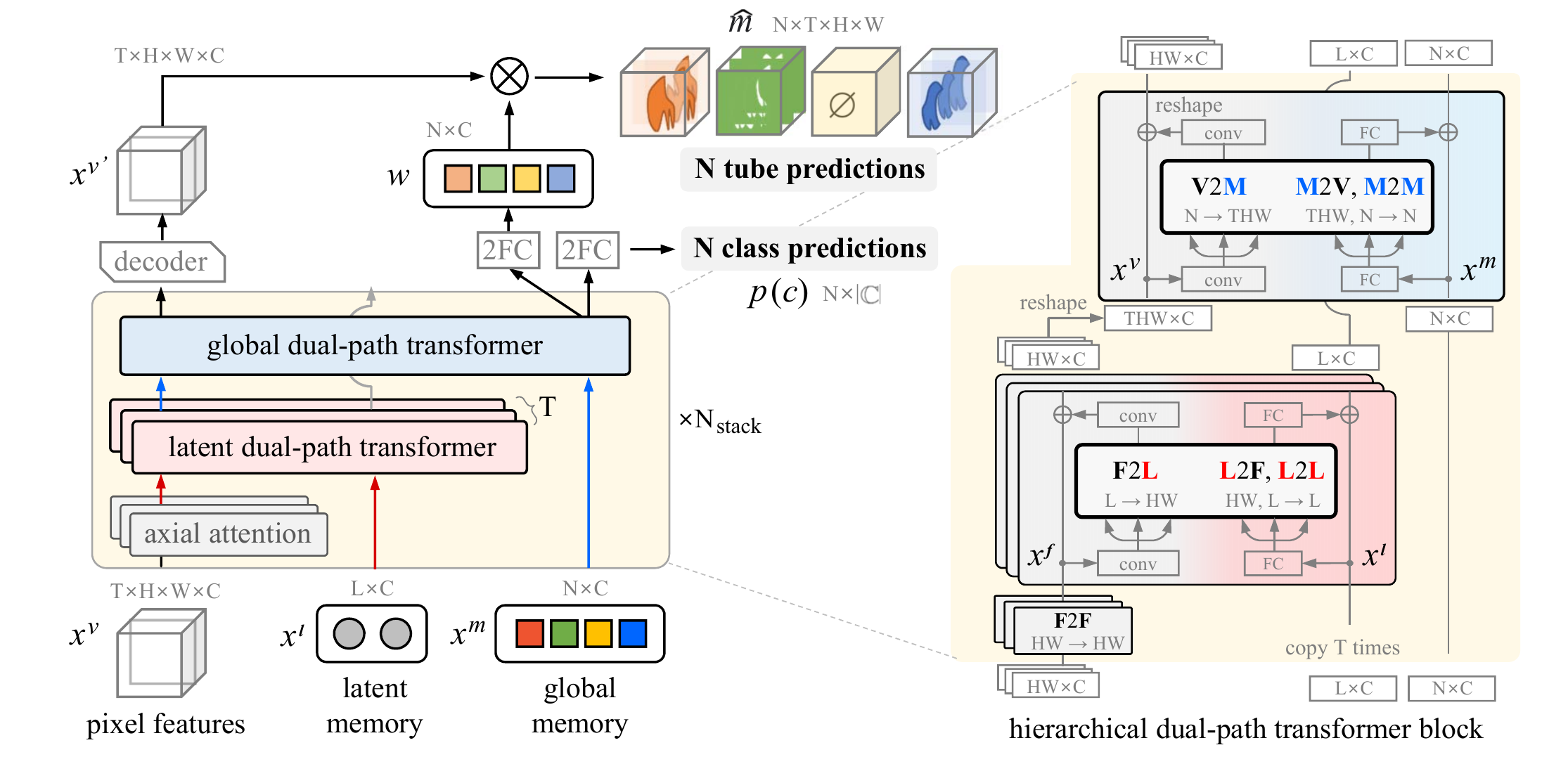}
\vspace{-2mm}
\caption{\textbf{TubeFormer-DeepLab architecture overview.} TubeFormer-DeepLab extends the mask transformer~\cite{wang2021max} to generate a set of pairs, each containing a class prediction $p(c)$ and a tube embedding vector $w$. The tube embedding vector, multiplied by the video pixel embedding features $x^{v'}$ obtained by a convolutional network, yields the tube prediction $\hat{m}$. We introduce a hierarchical structure with the latent dual-path transformer block that is in charge of passing messages between frame-level features $x^f$ and a latent memory $x^l$, followed by the global dual-path transformer block that learns the attentions between video-clip features $x^v$ and a global memory $x^m$. 
}
\label{fig:overview}
\vspace{-3mm}
\end{figure*}

We first introduce TubeFormer-DeepLab-Simple, our video-level baseline, which will be improved by our proposed latent dual-path transformer, resulting in the final TubeFormer-DeepLab.

\paragraph{TubeFormer-DeepLab-Simple.} \quad
We adopt the per-clip pipeline which takes a video clip and outputs clip-level results. Inspired by~\cite{wang2021max}, our TubeFormer-DeepLab-Simple integrates a CNN backbone and a global memory feature in a dual-path architecture, \ie, global dual-path transformer.

Given an input video clip ${v}$, the CNN backbone processes the input frames independently, and generates pixel features $x^v \in \mathbb{R}^{T\times H\times W\times C}$, where $C$ is channels. The pixel self-attention is performed at the frame level (frame-to-frame, F2F) via an axial-attention block~\cite{wang2020axial}.

Afterwards, the global dual-path transformer operates in a \textit{per-clip} manner, taking the flattened video pixel features $x^v \in \mathbb{R}^{THW\times C}$ and a 1D global memory $x^m \in \mathbb{R}^{N \times C}$ of length $N$ (\ie the size of the prediction set). Passing through the global dual-path transformer, we expect three attentions: (1) memory-to-video (M2V) attention (in which the video features encode per-clip information to the memory feature), (2) memory-to-memory (M2M) self-attention, and (3)  video-to-memory (V2M) attention (in which the video pixel features refine themselves by receiving tube-level information gathered in the global memory). The global dual-path transformer blocks can be stacked multiple times at any layers of the network. 

On top of the global memory, there are two output heads: a segmentation head and a class head, each composed of two Fully-Connected (FC) layers. The global memory of size $N$ is independently passed to the two heads, resulting in $N$ unique tube embeddings $w \in \mathbb{R}^{N\times C}$ and $N$ corresponding class predictions $p(c) \in \mathbb{R}^{N\times |\mathbb{C}|}$. Note that the possible classes $\mathbb{C} \ni c$ include ``none" category $\varnothing$ in case the embedding does not correspond to any region in a clip. Our video tube prediction $\hat{m}$ is computed in one shot as a dot-product between the decoded video pixel features $x^{v'}$ and the tube embeddings $w$:
\begin{equation}
\label{eq:output}
\hat{m} = \text{softmax}_N (x^{v'} \cdot w) \in \mathbb{R}^{N\times T\times H\times W}.
\end{equation}
The final video-clip segmentation $\{\hy_i\}_{i=1}^N = \{(\hatm_i, \hp_i(c))\}_{i=1}^N$ can be obtained by combining $N$ binary video tubes with their corresponding class predictions.

\paragraph{TubeFormer-DeepLab with Latent Dual-Path Transformer.} \quad
Modeling long-range interactions in \textit{video-clip} (\ie, multi-frames) features is especially difficult, when dealing with high-resolution inputs or a large number of input frames. To both alleviate the issue and facilitate the attention learning, we propose a hierarchical structure, which allows two levels of attention mechanisms: frame-level, followed by video-level. Note the video-level attention is performed by the aforementioned global dual-path transformer.

Prior to the global dual-path transformer, we introduce a new \textbf{latent} dual-path transformer block in charge of passing messages between \textit{frame-level} features and a \textbf{latent} memory.
It processes individual video frames in parallel (batchwise).
Our latent memory is inspired by the graphical models with latent representations~\cite{koller2009probabilistic,zhang2019latentgnn,hwang2021video}, allowing a low-rank representation for the graph affinity of high complexity. Concurrent with IFC~\cite{hwang2021video}, we discovered that latent features facilitate attention learning. However, we deployed them in a different framework (e.g., dual-path transformer and no cross-frame communication). 

Specifically, the initial latent memory $x^l \in \mathbb{R}^{L\times C}$ is copied per frame and paired with each frame's features $x^f \in \mathbb{R}^{HW\times C}$ (flattened) to construct the input. Passing through the latent dual-path transformer, the latent memory first collects messages from frame features via latent-to-frame (L2F) attention, and perform latent-to-latent (L2L) self-attention among themselves. Afterwards, the per-frame knowledge from the latent memory is propagated back to the frame features via frame-to-latent (F2L) attention. Note the latent memory features are trainable parameters like the global memory features. However, they are only deployed in the latent space (\ie, intermediate layers) and will not be used in the final output layers.

As shown in~\figref{fig:overview}, our hierarchical dual-path transformer blocks consist of a series of one axial-attention block, the latent dual-path transformer, and the global dual-path transformer. The stacking of multiple blocks will alternate the latent and the global communications, allowing the pixel features to refine themselves by attending to both frame-level and video-level memory, and vice versa. This in turn enriches the features of all three paths: pixel-, latent-memory and global-memory paths, and enables learning more comprehensive representations of the given video clip. 

\paragraph{Global memory with split thing and stuff.}\quad
To further improve the segmentation quality, we propose to split the global memory into two sets: thing-specific and stuff-specific global memory.
Originally, the global memory in ~\cite{wang2021max} deals with thing masks and stuff masks in a unified manner.
However, the design ignores the natural difference between them --- There could be multiple instances of the same thing class in an image, but at most one mask is allowed for each stuff class.
We thus allocate the last $|\mathbb{C}^\text{stuff}|$ out of $N$ elements in the global memory specifically for predicting stuff classes.
The ordering is enforced by assigning the stuff-specific global memory to the ground truth stuff classes, instead of including them in the bipartite matching.

\subsection{Training Strategy}

\paragraph{VPQ-style loss.}\quad
To train TubeFormer-DeepLab for various video segmentation tasks in a unified manner, we adopt a VPQ-style loss that directly optimizes the set of class-labeled tubes. Similar to the image-level PQ-style loss~\cite{wang2021max}, we draw inspiration from \textit{video panoptic quality} (VPQ)~\cite{kim2020video} and approximately optimize VPQ within a video clip.

To start with, a VPQ-style similarity metric between a class-labeled ground truth tube $y_i$ = $(m_i, c_i)$ and a predicted tube $\hy_j$ = $(\hatm_j, \hp_j(c))$ can be defined as: ${\rm sim} (y_i, \hy_j)$ = $\hp_j(c_i) \times {\rm Dice}(m_i, \hatm_j),$
where $\hp_j(c_i)\in[0, 1]$ denotes the probability of predicting the correct tube class $c_i$ and ${\rm Dice}(m_i, \hatm_j) \in [0, 1]$ measures the Dice coefficient between a predicted tube $\hatm_j$ and a ground truth tube $m_i$.

We match the predicted tubes to the ground truth tubes, and optimize the predictions by maximizing the total VPQ-style similarity. The
implementation details follow the PQ-style loss in~\cite{wang2021max}.
In addition, we generalize the auxiliary losses used in~\cite{wang2021max} to video clips, resulting in a tube-ID cross entropy loss, a video semantic segmentation loss, and a video instance discrimination loss.

\paragraph{Shared semantic and panoptic prediction.}\quad
Originally, the auxiliary semantic segmentation loss in~\cite{wang2021max} is applied to the backbone feature with a {\it separate} semantic decoder. Instead, we propose to apply the loss directly to the decoded video pixel features $x^{v'}$ (\cf~\equref{eq:output}) with a linear layer, which learns better features for segmentation.

\paragraph{Temporal consistency loss.}\quad
The VPQ-style loss benefits the learning of spatial-temporal consistency within an input clip. To further achieve the {\it clip-to-clip} consistency over a longer video, we propose to use a temporal consistency loss applied between clips. Specifically, we minimize the distance between the $N$ tube logits predicted from the overlapping frames of two clips. We use $L1$ loss for the consistency metric. The loss is back-propagated through the dot-product of the pixel features and $N$ global memory features, affecting both pixel and global memory paths. TubeFormer-DeepLab thereby achieves implicit multi-clip consistency, which makes our training objective symmetrical to the whole-video inference pipeline (\secref{sec:inference}).

\paragraph{Clip-level copy-paste.} \quad
Additionally, we propose a simple and effective data augmentation policy by extending the image-level thing-specific copy-paste~\cite{fang2019instaboost, ghiasi2021simple}. Our augmentation method, named clip-paste (clip-level copy-paste), randomly pastes either `thing' or `stuff' (or both) region tubes from a video clip to the target video clip.
We use clip-paste with a probability of 0.5.

\paragraph{Depth prediction branch.} \quad
To grant TubeFormer-DeepLab the ability to perform monocular depth estimation, we add a small depth prediction module (\ie, ASPP~\cite{chen2018deeplabv2} and DeepLabv3+ lightweight decoder~\cite{deeplabv3plus2018}) on top of the CNN {\it backbone} features $x^{v}$. Note that we found the performance slightly degrades if we add the depth prediction to the decoded video pixel features $x^{v'}$, indicating that it is not beneficial to share depth estimation with segmentation prediction in our case.
We apply Sigmoid to constrain the depth prediction to the range (0, 1), and then multiply it by the maximum depth.
Following ~\cite{qiao2021vip}, we use the combination of scale invariant logarithmic error~\cite{eigen2014depth} and relative squared error~\cite{geiger2012we} as the training loss.
The depth loss weight is set to $100$ when jointly trained with the other losses.

\subsection{Inference Strategy}
\label{sec:inference}

\paragraph{Clip-level inference.}\quad
The clip-level segmentation is inferred by simply performing argmax twice. Specifically, a class label is predicted for each tube: ${\hatc_i = \argmax_c \hp_i(c) \,.}$ And then, a tube-ID $\hz_{t,h,w}$ is assigned \textbf{per-pixel}: ${\hz_{t,h,w} = \argmax_i \hatm_{i,t,h,w} \,.}$ In practice, our inference sets tube-IDs with class confidence below 0.7 to void.

For video instance segmentation, we also explore \textbf{per-mask} assignment scheme~\cite{cheng2021per,qihang2022cmt}, which treats the prediction of each object query as one object mask proposal.

\paragraph{Video-level inference.}\quad
At the clip level, TubeFormer-DeepLab outputs temporally consistent results for $T$ video frames.
To obtain the video-level prediction, we perform clip-level inference for every $T$ consecutive frames with $T-1$ overlapping frames (\ie, we move along the temporal axis by only one frame at each inference step). The clip-level results are then stitched together by matching tubes in the overlapping frames based on their IoUs, similar to ~\cite{qiao2021vip}.

\section{Experimental Results}
\label{sec:experiments}

Our proposed TubeFormer-DeepLab is a general video segmentation model. To demonstrate its effectiveness, we conduct experiments on KITTI-STEP~\cite{Weber2021NEURIPSDATA}, VIPSeg~\cite{miao2022large}, VSPW~\cite{miao2021vspw}, YouTube-VIS~\cite{Yang19ICCV}, SemKITTI-DVPS~\cite{qiao2021vip} for Video Panoptic Segmentation (VPS), Video Semantic Segmentation (VSS), Video Instance Segmentation (VIS), and Depth-aware Video Panoptic Segmentation (DVPS), respectively.

\subsection{Datasets}

{\bf KITTI-STEP}~\cite{Weber2021NEURIPSDATA} is a new video panoptic segmentation dataset that additionally annotates semantic segmentation for KITTI-MOTS ~\cite{voigtlaender2019mots}.
It contains 19 semantic classes (similar to Cityscapes~\cite{cordts2016cityscapes}), among which two classes (`pedestrians' and `cars') come with tracking IDs. For evaluation, KITTI-STEP adopts STQ~\cite{Weber2021NEURIPSDATA} (segmentation and tracking quality), which is the geometric mean of SQ (segmentation quality) and AQ (association quality).

{\bf VIPSeg}~\cite{miao2022large} is also a new video panoptic segmentation dataset for diverse in-the-wild scenarios.
It contains 124 semantic classes (58 `thing' and 66 `stuff' classes) with 3536 videos, where each video spans 3 to 10 seconds.

{\bf VSPW}~\cite{miao2021vspw} is a recent large-scale video semantic segmentation dataset, containing 124 semantic classes.
VSPW adopts mIoU as the  evaluation metric.

{\bf YouTube-VIS}~\cite{Yang19ICCV} contains two versions for video instance segmentation; The YouTube-VIS-2019 contains 40 semantic classes and the YouTube-VIS-2021 is an improved version with higher number of instances and videos. Youtube-VIS adopts track mAP for evaluation.

{\bf SemKITTI-DVPS}~\cite{qiao2021vip} is a new dataset for depth-aware video panoptic segmentation, which is obtained by projecting the 3D point cloud panoptic annotations of SemanticKITTI~\cite{behley2019semantickitti} to 2D image planes. It contains 19 classes, among which 8 are annotated with tracking IDs.
For evaluation, SemKITTI-DVPS uses DSTQ (depth-aware STQ), which considers depth inlier metric~\cite{eigen2014depth} in addition to STQ.

\subsection{Implementation Details}
TubeFormer-DeepLab builds upon MaX-DeepLab~\cite{wang2021max} with the official codebase~\cite{deeplab2_2021}.
The hyper-parameters mostly follow the settings of ~\cite{wang2021max}.
Unless specified, we use their small model MaX-DeepLab-S, which augments ResNet-50~\cite{he2016deep} with axial-attention blocks~\cite{wang2020axial} in the last two stages (\ie, stage-4 and stage-5). We also experiment with scaling up the backbone ~\cite{swidernet_2020} by stacking the axial-attention blocks in stage-4 by {\bf $n$} times, and refer them as TubeFormer-DeepLab-B{\bf $n$} in the experiments.
For VPS, we pretrain the models on Cityscapes~\cite{cordts2016cityscapes} and COCO~\cite{lin2014microsoft}, while for other experiments, we only pretrain on COCO.
The pretraining procedure is similar to prior works~\cite{Weber2021NEURIPSDATA,bertasius2020classifying,he2021exploiting}.
%We first pretrain the model on COCO~\cite{lin2014microsoft}. The pretraining runs 200K iterations with a batch size of 32 (54 epochs).
Using the pretrained weights, TubeFormer-DeepLab is trained on the target datasets using a batch size of 16, with $T$ = 2 for all datasets except $T$ = 5 for YouTube-VIS dataset. We use the global memory size $N$ = 128 (\ie, output size), latent memory size $L$ = 16, and $C$ = 128 channels.
We use `TF-DL' to denote TubeFormer-DeepLab in the results.

\subsection{Main Results}

\paragraph{[VPS]}\quad We evaluate TubeFormer-DeepLab on the challenging video panoptic segmentation dataset, KITTI-STEP~\cite{Weber2021NEURIPSDATA} in \tabref{tab:kitti_step}. Our model achieves state-of-the-art performance with 65.25 STQ (70.27 SQ and 60.59 AQ). Among {\it single unified} approaches, our model ranks first, significantly outperforming the published baseline Motion-DeepLab~\cite{Weber2021NEURIPSDATA} by {\bf +13.1} STQ. Our model performs comparably with the challenge winning methods~\cite{zhang2021u3dmolts,lu2021robust} without exploiting extra 3D object formulation, depth information, or pseudo labels, and even without the employment of separate and ensemble methods for tracking and segmentation. Nevertheless, our model delivers the best segmentation quality (70.27 SQ), showcasing our TubeFormer-DeepLab's segmentation ability.

We further evaluate TubeFormer-DeepLab on the recent video panoptic segmentation dataset, VIPSeg~\cite{miao2022large} in~\tabref{tab:vipseg_test}.
Our method outperforms Clip-PanoFCN~\cite{miao2022large} (which built on top of Panopitc FCN~\cite{li2021fully}) by {\bf +13.6} STQ and {\bf +3.9} VPQ on the test set.

\begin{table}[t]
\centering
\small
\tablestyle{8pt}{1.1}
\begin{tabular}{l|c|ccc}
method                              & rank  & STQ       & SQ        & AQ        \\
\shline
Motion-DeepLab~\cite{Weber2021NEURIPSDATA} & 7     & 52.19     & 59.81     & 45.55     \\
\hline
\multicolumn{5}{c}{ICCV 2021 challenge entries}\\
\hline
HybridTracker                       & 6     & 54.99     & 55.54     & 55.54     \\
slain                               & 5     & 57.87     & 60.71     & 55.16     \\
EffPs\_MM                           & 4     & 62.93     & 64.41     & 61.49     \\
REPEAT~\cite{lu2021robust}          & 2     & 67.13     & 68.49     & 65.81     \\
UW\_IPL/ETRI\_AIRL~\cite{zhang2021u3dmolts}   
                                    & 1     & \textbf{67.55}     & 64.04     & \textbf{71.26}     \\
\hline
\hline
TF-DL-B3                            & 3     & 65.25     & \textbf{70.27}     & 60.59     \\
\end{tabular}
\vspace{-.8em}
\caption{\textbf{[VPS]} KITTI-STEP {\it test} set results. Ranking includes unpublished methods. The challenge winning entries~\cite{zhang2021u3dmolts,lu2021robust} adopt separate and ensemble methods for tracking and segmentation.
}
\label{tab:kitti_step}
\vspace{-1mm}
\end{table}

\begin{table}[t]
\centering
\small
\tablestyle{14pt}{1.1}
\begin{tabular}{l|ccc}
method                              &  STQ & VPQ\\
\shline
\multicolumn{3}{l}{\textit{val set}}\\
Clip-PanoFCN~\cite{miao2022large}   & 31.5 & 22.9 \\
\hline
TF-DL-B1                            & 39.8 & 29.2 \\
TF-DL-B3                            & 41.5 & 31.2 \\
\hline \hline
\multicolumn{3}{l}{\textit{test set}}\\
Clip-PanoFCN~\cite{miao2022large}   & 25.0 & 22.9 \\
\hline
TF-DL-B3                            & 38.6 & 26.8 \\
\end{tabular}
\vspace{-.8em}
\caption{\textbf{[VPS]} VIPSeg {\it val} and {\it test} set results, using the latest test server at \url{https://codalab.lisn.upsaclay.fr/competitions/9743} 
}
\label{tab:vipseg_test}
\vspace{-1mm}
\end{table}

\paragraph{[VSS]}\quad
We assess TubeFormer-DeepLab on the video semantic segmentation dataset, VSPW~\cite{miao2021vspw}. We show the single-model single-scale results on {\it val} set in \tabref{tab:vspw_val}. 
In the table, TubeFormer-DeepLab outperforms all competing methods, which are based on state-of-the-art backbones (BEiT~\cite{bao2021beit}, Swin-L~\cite{liu2021swin}) and decoders (OCRNet~\cite{yuan2020object}, UperNet~\cite{xiao2018unified}). 
\tabref{tab:vspw_test} shows the {\it test} set results. Our single-model TubeFormer-DeepLab achieves competitive results (rank 4 out of 17) with the ICCV 2021 challenge winners, while not employing model ensembles, multi-scale inference, and pseudo labels.
Finally, we attain a better {\bf +21} mIoU than the published work TCB~\cite{miao2021vspw} on the {\it test set}.
As shown in the bottom of~\tabref{tab:vspw_test}, we also include the new test set results using the latest test server.

\begin{table}[t]
\centering
\small
\tablestyle{3pt}{1.1}
\begin{tabular}{l|ccc}
method                      & mIoU      & VC8       & VC16       \\
\shline
TCB~\cite{miao2021vspw}     & 37.82     & 87.86     & 83.99     \\
\hline
\multicolumn{4}{c}{ICCV 2021 challenge entries}\\
\hline
BetterThing~\cite{chen2021semantic}
                            & 57.89     & -         & -         \\
CharlesBLWX~\cite{jin2021memory} 
                            & 61.44     & -         & -         \\
jjRain~\cite{he2021exploiting}  
                            & 59.30     & 90.07     & 86.87     \\
\hline
\hline
TF-DL-B4                    & \bf{63.16} & \bf{92.08}  & \bf{87.95} \\
\end{tabular}
\vspace{-.8em}
\caption{\textbf{[VSS]} VSPW \textit{val} set results. Comparison includes published and unpublished methods. 
}
\label{tab:vspw_val}
\end{table}

\begin{table}[t]
\centering
\small
\tablestyle{3pt}{1.1}
\begin{tabular}{l|c|ccc|ccc}
method     & rank      & ens.    & m.s.      & pseudo        & mIoU      & VC8      & VC16       \\
\shline
\multicolumn{8}{l}{\textit{old codalab}}\\
TCB~\cite{miao2021vspw}     & 13        &       &        &               & 35.62     & 86.21    & 81.90      \\
\hline
\multicolumn{8}{c}{ICCV 2021 challenge entries}\\
\hline
BetterThing~\cite{chen2021semantic}
                            & 3        & \checkmark    & \checkmark   &             & 57.35     & 93.28     & 90.56      \\
CharlesBLWX~\cite{jin2021memory}          
                            & 2        & \checkmark    & \checkmark  &                & 57.44     & 91.29     & 87.70     \\
jjRain~\cite{he2021exploiting}         
                            & 1        & \checkmark    & \checkmark & \checkmark     & {\bf 58.85}     & {\bf 94.77}     & {\bf 92.59}     \\
\hline

TF-DL-B4                     & 4             &           &    &             & 56.64     & 90.16     & 86.38 \\
\hline \hline
\multicolumn{8}{l}{\textit{new codalab}}\\
TCB~\cite{miao2021vspw}     & & & & & 32.58     & 79.46     & 73.23   \\
\hline
TF-DL-B4 & & & & & \textbf{52.99} & \textbf{90.16} & \textbf{86.38} \\

\end{tabular}
\vspace{-.8em}
\caption{\textbf{[VSS]} VSPW \textit{test} set results. Ranking includes published and unpublished methods. Some methods use model ensembles, multi-scale inference, or teacher-student pseudo labeling strategy to boost performance on test set.
In the bottom rows, we also include the new test set results, using the latest test server at \url{https://codalab.lisn.upsaclay.fr/competitions/7869}
}
\label{tab:vspw_test}
\vspace{-2mm}
\end{table}

\paragraph{[VIS]}\quad
We show that TubeFormer-DeepLab is sufficiently general to solve instance-level video segmentation in a unified manner. The same model, loss, and training procedure is seamlessly applied by treating the background region as a single `stuff' class. At testing, we explore both per-pixel and per-mask argmax for tube ID assignment (\secref{sec:inference}).  

\tabref{tab:yt_vis_19} and \ref{tab:yt_vis_21} show the comparison with the state-of-the-art methods on YouTube-VIS 2019 and 2021 datasets~\cite{Yang19ICCV}. Note that TubeFormer-DeepLab predicts a single unique mask per object, while other methods often generate multiple overlapping masks, which are favored by the AP metric. Among end-to-end methods, our TubeFormer-DeepLab-B4 outperforms VisTR~\cite{wang2021end} by {\bf +7.4}, and IFC~\cite{hwang2021video} by {\bf +2.9} AP. Our model with $T = 5$ sets the highest scores among methods that employ a small value of $T$.
Also, our gains in AR\textsubscript{1} are significant, indicating the benefit of TubeFormer-DeepLab in the non-overlapping segmentation scenario.

Our model performs comparably to Seq Mask R-CNN~\cite{lin2021video}. We point out that TubeFormer-DeepLab is an end-to-end near-online method, while Seq Mask R-CNN relies on STM~\cite{oh2019video}-like structure to propagate mask proposals through the whole sequence, and thus is offline ($T$=36).

\paragraph{[DVPS]}\quad We evaluate TubeFormer-DeepLab on the SemKITTI-DVPS dataset~\cite{qiao2021vip} for depth-aware video panoptic segmentation. \tabref{tab:semkitti_dvps} shows the {\it test} set results. Adding a depth prediction branch to the same exact TubeFormer-DeepLab used for KITTI-STEP outperforms ViP-DeepLab~\cite{qiao2021vip} by {\bf +3.4} DSTQ and achieves the new state-of-the-art of 67.0 DSTQ.

\begin{table}[!t]
\centering
\small
\tablestyle{4.0pt}{1.1}
\begin{tabular}{l|c|ccc|cc}
        method         & T  & AP        & AP\textsubscript{50} & AP\textsubscript{75} & AR\textsubscript{1}  & AR\textsubscript{10}     \\
\shline
        MaskTrack~\cite{Yang19ICCV}    & 2  & 31.8      & 53.0      & 33.6      & 33.2      & 37.6       \\
        SipMask~\cite{cao2020sipmask}  & 2  & 33.7      & 54.1      & 35.8      & 35.4      & 40.1        \\
        STEm-Seg~\cite{athar2020stem}  & 8  & 34.6      & 55.8      & 37.9      & 34.4      & 41.6        \\
        CrossVIS~\cite{yang2021crossover} 
                                        & 2  & 36.6      & 57.3      & 39.7      & 36.0      & 42.0       \\
        MaskProp~\cite{bertasius2020classifying}  
                                        & 13  & 46.6      & -         & 51.2      & 44.0         & 52.6 \\
        %                                & 13  & 42.5      & -         & 45.6      & -         & -          \\
        Seq Mask R-CNN~\cite{lin2021video}  
                                        & 36  & {\bf 47.6}      & {\bf 71.6}    & 51.8      & 46.3         & 56.0 \\
        
        VisTR~\cite{wang2021end}       & 36 & 40.1      & 64.0      & 45.0      & 38.3      & 44.9       \\
        IFC~\cite{hwang2021video}      & 36 & 44.6      & 69.2      & 49.5      & 44.0      & 52.1       \\

\hline
        \hline
        TF-DL-B4 (per-pixel)                       & 5  & 45.4      & 66.6      & 48.8      & {48.3} & {56.9}    \\
        {TF-DL-B4} (per-mask)                      & 5  & {47.5}  & 68.7      & \textbf{52.1}   & \textbf{50.2} & \textbf{59.0}    \\
\end{tabular}
\vspace{-.8em}
\caption{\textbf{[VIS]} YouTube-VIS-2019 {\it val} set results.
}
\label{tab:yt_vis_19}
\vspace{-1mm}
\end{table}

\begin{table}[!t]
\centering
\small
\tablestyle{4pt}{1.1}
\begin{tabular}{l|c|ccc|cc}
                    method          & T  & AP        & AP\textsubscript{50} & AP\textsubscript{75}  & AR\textsubscript{1}  & AR\textsubscript{10}     \\
\shline
        MaskTrack~\cite{Yang19ICCV}    & 2  & 28.6      & 48.9      & 29.6   & - & -   \\
        SipMask~\cite{cao2020sipmask}  & 2$^\dagger$  & 31.7      & 52.5      & 34.0   & - & -    \\
        CrossVIS~\cite{yang2021crossover} 
                                        & 2$^\dagger$  & 34.2      & 54.4      & 37.9  & - & -     \\
        IFC~\cite{hwang2021video}      & 36$^\dagger$  & 36.8      & 57.9      & 39.3  & - & -     \\

\hline
        TF-DL-B4 (per-mask)           & 5  & {\bf 41.2}      & {\bf 60.4}      & {\bf 44.7}      &\textbf{40.4} & \textbf{54.0}    \\
\end{tabular}
\vspace{-.8em}
\caption{\textbf{[VIS]} YouTube-VIS-2021 {\it val} set results. $^\dagger$: T inferred from their Youtube-VIS-2019 settings. }
\label{tab:yt_vis_21}
\vspace{-1mm}
\end{table}

\begin{table}[!t]
\centering
\small
\tablestyle{8pt}{1.1}
\begin{tabular}{l|c|c}
method                          & rank  & DSTQ \\
\shline
ViP-DeepLab~\cite{qiao2021vip}  & 3     & 63.36 \\
\hline
\multicolumn{3}{c}{ICCV 2021 challenge entries}\\
\hline
rl\_lab                         & 5     & 54.77 \\
ywang26                         & 4     & 55.99 \\
HarborY~\cite{li2021polyphonicFormer}  & 2     & 63.63 \\
\hline
\hline
TF-DL-B4                       & 1     & \textbf{67.00} \\
\end{tabular}
\vspace{-.8em}
\caption{\small{\textbf{[DVPS]} SemKITTI-DVPS {\it test} set results. Ranking includes published and unpublished methods.}}
\label{tab:semkitti_dvps}
\vspace{-1mm}
\end{table}

\begin{figure*}[t]
\centering
\includegraphics[width=0.95\linewidth]{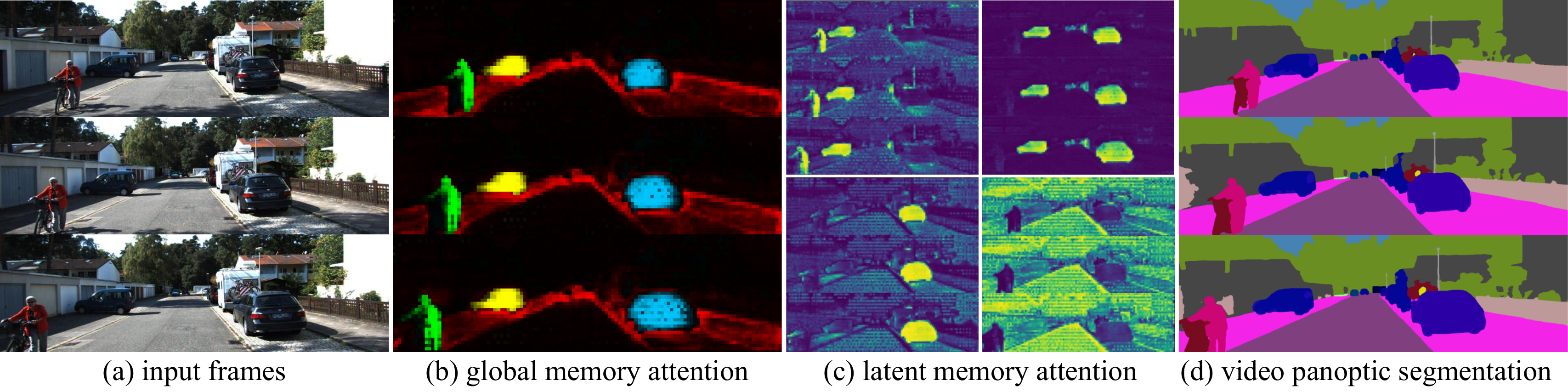}
\vspace{-1mm}
\caption{\small \textbf{Visualization} on KITTI-STEP sequence. From left to right: input frames ($T = 3$), global memory attention, latent memory attention, and video panoptic segmentation results. The global memory attention is selected for predicted tube regions of interest: a {\color[HTML]{009901}\textbf{pedestrian}} and two cars ({\color[HTML]{F8A102} \textbf{left}}, {\color[HTML]{3166FF}\textbf{right}}) on the {\color[HTML]{FE0000}\textbf{sidewalk}}, and the latent memory attention is selected for 4 (out of $L = 16$)  latent memory.
}
\label{fig:visualization}
\end{figure*}

\subsection{Ablation Studies}
We provide ablation studies on the KITTI-STEP \textit{val} set~\cite{Weber2021NEURIPSDATA}.
To compensate for the training noise, we report the mean of three runs for every ablation study.

\paragraph{Hierarchical dual-path transformer.}\quad In \tabref{tab:ablations:attention_type}, we verify that the gains demonstrated by TubeFormer-DeepLab come from the proposed hierarchical dual-path transformer. Note that our baseline method (TubeFormer-DeepLab-Simple) already uses the axial attentions and the global memory. Introducing the new latent memory and its communication with the video-frame features (F-L attention: L2F, L2L, and F2L) brings a large improvement of +1.7 STQ. We also ablate adding attentions between the global memory and the latent memory (M2L and L2M), which show no improvements. This suggests the frame-latent (F-L) attention is sufficient to build effective hierarchical attentions between the latent and the global dual-path transformers. We also ablate different latent memory size, and set the default size $L$ to be 16.

\paragraph{Training strategy.}\quad In addition, \tabref{tab:ablations:others} shows that the proposed temporal consistency loss helps TubeFormer-DeepLab to learn clip-to-clip consistency, and improves the inference on longer videos than the training clip length ($T$), as demonstrated by +0.5 STQ gain. The proposed clip-level copy-paste (clip-paste) augments more training samples for tube-level segmentation, and further improves by +0.9 STQ. 

\begin{table}[t]
\vspace{-2mm}
\centering
% subfloat a - Attention types
\subfloat[{\textbf{Varying transformer attention types}. Frame-latent (\textbf{F-L}) attention is introduced in the proposed latent dual-path transformer, and includes latent-to-frame, latent-to-latent, and frame-to-latent attentions. We also ablate  memory-to-latent (\textbf{M2L}) and latent-to-memory (\textbf{L2M}) attentions, and different latent memory size $L$.}]{
\label{tab:ablations:attention_type}
\tablestyle{3pt}{1.1}
\begin{tabular}{l|c|ccc|ccc}
method      & $L$ & F-L        & M2L       & L2M       & STQ       & SQ        & AQ         \\
\shline
TF-DL-Simple
            & 0 &            &           &           & 68.36      & 74.93    & 62.38     \\ 
\hline            
\multirow{5}{*}{\parbox{2.3cm}{TF-DL (with \\ {{\color[HTML]{FFFFFF}-}hierarchical\\ {\color[HTML]{FFFFFF}-}dual-path\\ {\color[HTML]{FFFFFF}-}transformer}) }}
            & \textbf{16} &\checkmark  &           &           & \bf{70.03} & \bf{76.83}	& \bf{63.83}  \\
            & 16 &\checkmark  &\checkmark &           &  69.63     & 76.05      & 63.75      \\
            & 16 &\checkmark  &           &\checkmark &  69.64     & 76.75      & 63.18       \\
            & 8  &\checkmark  &           &           & 69.39      & 75.74      & 63.54       \\
            & 32 &\checkmark  &           &           & 69.57      & 76.71      & 63.1        \\

\end{tabular}
}\vspace{1mm}

% subfloat b - TC, ClipPaste, Kalman
\subfloat[{Adding \textbf{temporal consistency loss} and \textbf{clip-level copy-paste}.}
]{
\label{tab:ablations:others}
\tablestyle{8pt}{1.1}
\begin{tabular}{lc|ccc}
method             &     & STQ       & SQ        & AQ         \\
\shline
TF-DL
                    &    & 70.03      & 76.83	& 63.83     \\ 
\hline            
+ temporal consistency  && 70.51      & \textbf{77.64}     & 64.04      \\
+ clip-paste            && \textbf{71.40}      & 76.82     & \textbf{66.36}      \\
% + Kalman filtering      & \bf{xx.xx} & \bf{xx.xx} & \bf{xx.xx}  \\
\end{tabular}
}\vspace{1mm}

% subfloat c - Scaling
\subfloat[{\textbf{Scaling} by stacking axial blocks in stage-4 of Axial-ResNet-50 by \textbf{n} times and \textbf{pretrain}ing on ImageNet-22K and COCO.}]{
\label{tab:ablations:scaling}
\tablestyle{7pt}{1.1}
\begin{tabular}{l|c|lc|ccc}
      & n         & INet    & COCO        & STQ        & SQ       & AQ     \\
\shline
B1    & 1         & 1k      &             & 70.03      & 76.83	  & 63.83   \\
B1    & 1         & 1k+22k  &             & 72.28      & 76.27    & 67.01  \\
B1    & 1         & 1k+22k  & \checkmark  & 73.19      & 78.11    & 68.58  \\
B3    & 3         & 1k+22k  & \checkmark  & {\bf 74.25} & {\bf 78.31} & {\bf 70.04}  \\
B4    & 4         & 1k+22k  & \checkmark  & 73.68      & 78.16    & 69.46  \\
\end{tabular}
}\vspace{1mm}

% subfloat d - Architectural improvements
\subfloat[{Ablating \textbf{architectural improvements}}]{
\label{tab:ablations:image_level}
\tablestyle{5.5pt}{1.1}
\begin{tabular}{l|ccc}
                                & STQ        & SQ       & AQ    \\
\shline
% hierarchical dual-path transformer             & \bf{70.03} & \bf{76.83}	& \bf{63.83}  \\
TF-DL                                  & \bf{70.03} & \bf{76.83}	& \bf{63.83}  \\
\hline
\textit{without} sharing semantic and panoptic & 68.95      & 75.83     & 62.70 \\
\textit{without} split thing and stuff memory  & 68.96	    & 75.77     & 62.76 \\
\end{tabular}
} 
% main caption
\vspace{-1mm}
\caption{Ablation studies on KITTI-STEP {\it val} set.}
\vspace{-2mm}
\label{tab:ablations}
\end{table}

\paragraph{Scaling.}\quad We study the scaling of TubeFormer-DeepLab in \tabref{tab:ablations:scaling}. Pretraining on ImageNet-22k dataset brings +1.6 STQ and adding COCO to the training further gives +1.9 STQ. We also explore scaling up the backbone by stacking the axial-attention blocks in stage-4 by $n$ times (TubeFormer-DeepLab-B\textit{n}). The increase of every $n$ will introduce +13M parameters. We notice increasing the stack from $n=1$ to $n=3$ improves the STQ from 73.19 to 74.25. Further scaling to $n=4$ starts to saturate, probably limited by the scale of KITTI-STEP dataset. We observe TubeFormer-DeepLab can further scale to $n=4$ on larger-scale datasets, where TubeFormer-DeepLab-B4 performed better on VSPW and YouTube-VIS datasets.

\paragraph{Architectural improvements.}\quad
We ablate our new architectural designs: (1) sharing the semantic and panoptic predictions, and (2) splitting the global memory for separate thing and stuff classes.
As shown in \tabref{tab:ablations:image_level}, we observe a performance drop of -1.1 STQ  by reverting the change of either (1) or (2) from TubeFormer-DeepLab.

\subsection{Visualization}
In \figref{fig:visualization}, we visualize how the proposed hierarchical dual-path transformer performs attention onto the input clip of three consecutive frames. We first visualize the global memory attention by selecting four output regions of interest from TubeFormer-DeepLab video panoptic prediction. We probe the attention weights between the four tube-specific global memory embeddings and all the pixels. We see the global memory attention is spatio-temporally well separated for individual thing or stuff tubes.

In addition, we select four latent memory indices and visualize their attention maps in \figref{fig:visualization}{\color[HTML]{FF0000}c}. We find that some latent memory learns to spatially specialize on certain areas (left \textit{vs} right side of the scene) or attends to semantically-similar regions (cars or backgrounds) to facilitate per-frame attention. With the hierarchical attentions made by the global and latent dual-path transformers, TubeFormer-DeepLab can be a successful \textit{tube} transformer.

Finally, we provide more visualizations for each video segmentation task in \secref{sec:visualization} and {\it video} prediction results at \href{https://youtu.be/twoJyHpkTbQ}{https://youtu.be/twoJyHpkTbQ}.

\section{More Experimental Results}

In this section, we provide more experimental results, comparing our methods with {\it published} works in detail. We do not include the {\it unpublished and concurrent} ICCV 2021 challenge entries, which usually adopt complicated pipelines, \eg, model ensembles, separate models for different sub-tasks (\eg, tracking, and segmentation), multi-scale inference, or pseudo labels. In the tables, we explicitly list the adopted backbones and decoders for a detailed comparison. We note that most of the state-of-the-art approaches for different video segmentation tasks have fundamentally diverged, while our proposed TubeFormer-DeepLab is a simple and unified system for general video segmentation tasks.

\paragraph{[VPS]} \quad
\tabref{tab:sup_kitti_step} summarizes our results on KITTI-STEP {\it val} set. As shown in the table, our TubeFormer-DeepLab-B1, employing ResNet-50~\cite{he2016deep} and axial-attention~\cite{wang2020axial}, significantly outperforms Motion-DeepLab~\cite{Weber2021NEURIPSDATA} (w/ ResNet-50, dual-ASPP~\cite{chen2017deeplabv3} and dual decoders~\cite{deeplabv3plus2018}) and VPSNet~\cite{kim2020video} (w/ ResNet-50, FPN~\cite{lin2017feature}, and Mask R-CNN~\cite{he2017mask} multi-head predictions) by {\bf +12} and {\bf +14} STQ, respectively. We also report the results in the VPQ metric~\cite{kim2020video} (another popular video panoptic segmentation metric). Similarly, our model performs better than Motion-DeepLab and VPSNet by {\bf +11.1} and {\bf +8.1} VPQ.

\paragraph{[VSS]} \quad
In \tabref{tab:sup_vspw_val}, we report our results on VSPW {\it val} set. As shown in the table, our TubeFormer-DeepLab-B1, employing ResNet-50 and axial-attention, significantly outperforms TCB~\cite{miao2021vspw} (w/  spatial-temporal OCRNet~\cite{yuan2020object} and a novel memory scheme) by {\bf +20.2} mIoU. Our TubeFormer-DeepLab-B1 also shows better results in terms of VC8 and VC16 (another video semantic segmentation metrics proposed in~\cite{miao2021vspw}).

\paragraph{[VIS]} \quad
\tabref{tab:sup_yt_vis_19} summarizes our results on Youtube-VIS-2019 {\it val} set, along with several state-of-the-art methods.

Among the methods that predict non-overlapping segmentation, our TubeFormer-DeepLab-B1 (per-pixel), employing ResNet-50 and axial-attention, outperforms STEm-Seg~\cite{athar2020stem} (using ResNet-50, FPN, and their novel 3D convolution-based TSE decoder with multi-head predictions) by {\bf +5.8} AP. Our TubeFormer-DeepLab-B1 (per-pixel) is also better than STEm-Seg with ResNet-101 backbone by {\bf +1.8} AP. If we also increase our backbone capacity, our TubeFormer-DeepLab-B4 (per-pixel) performs better than STEm-Seg w/ ResNet-101 by {\bf +10.8} AP.

Our TubeFormer-DeepLab-B1 (per-pixel) performs worse than other state-of-the-art methods, including MaskProp~\cite{bertasius2020classifying}, Seq Mask R-CNN~\cite{lin2021video}, and the concurrent work IFC~\cite{hwang2021video}, since our per-pixel inference scheme generates non-overlapping predictions (\ie, only one prediction for each pixel in the final output), which is disfavored by the track AP metric. To bridge the gap, we adopt the mask-wise merging scheme (denoted as per-mask)~\cite{cheng2021per,qihang2022cmt}, where each object query generates a mask proposal. The per-mask scheme significantly improves over the per-pixel scheme by more than 2 AP in the TubeFormer-DeepLab framework. Our large model TubeFormer-DeepLab-B4 with per-mask scheme outperforms MaskProp, VisTR, and IFC, and performs comparably with the best model Seq Mask R-CNN, which relies on STM~\cite{oh2019video}-like structure to propagate mask proposals through the whole sequence.

Notably, our model yields the best $AR_1$ and $AR_{10}$ ({\bf +3.9} and {\bf +3.0} AR better than the second best Seq Mask-RCNN method, respectively), demonstrating the high segmentation quality in our predictions. Also, TubeFormer-DeepLab employs a smaller clip value ($T=5$), while other state-of-the-art proposal-based approaches use a large value of clip ($T=13$ or $36$).

\begin{table*}[!t]
\centering
\small
\tablestyle{6pt}{1.1}
\begin{tabular}{l|l|l|ccc|c}
method      & backbone & decoder  & STQ       & SQ        & AQ        & VPQ   \\
\shline
Motion-DeepLab~\cite{Weber2021NEURIPSDATA}   
            & ResNet-50 + dual ASPP~\cite{chen2017deeplabv3}  & dual DeepLabv3+ decoder~\cite{deeplabv3plus2018} w/ multi-heads & 58.0      & 67.0      & 51.0      & 40.0  \\
VPSNet~\cite{kim2020video}                   
            & ResNet-50 + FPN~\cite{lin2017feature}  & Mask R-CNN~\cite{he2017mask} style multi-heads & 56.0      & 61.0      & 52.0      & 43.0  \\
\hline\hline

TF-DL-B1 & ResNet-50 + axial-attention~\cite{wang2020axial}$\dagger$ & tube-transformer & \textbf{70.0} & \textbf{76.8} & \textbf{63.8} & \textbf{51.1} \\
\end{tabular}
\vspace{-.8em}
\caption{\textbf{[VPS]} KITTI-STEP {\it val} set results. $\dagger$: Axial attention blocks~\cite{wang2020axial} are used in the last two stages.
}
\label{tab:sup_kitti_step}
\end{table*}

\begin{table*}[!t]
\centering
\small
\tablestyle{7pt}{1.1}
\begin{tabular}{l|l|l|ccc}
method      & backbone  & decoder   & mIoU      & VC8       & VC16       \\
\shline
TCB~\cite{miao2021vspw}     
            & ResNet-101   & spatial-temporal OCRNet~\cite{yuan2020object} + memory aggregation & 37.8     & 87.9     & 84.0     \\
\hline\hline
TF-DL-B1  
    & ResNet-50 + axial-attention~\cite{wang2020axial}$\dagger$   & tube-transformer & \bf{58.0} & \bf{90.1}  & \bf{86.8} \\
\end{tabular}
\vspace{-.8em}
\caption{\textbf{[VSS]} VSPW \textit{val} set results. $\dagger$: Axial attention blocks~\cite{wang2020axial} are used in the last two stages.}
\label{tab:sup_vspw_val}
\end{table*}

\begin{table*}[!t]
\centering
\small
\tablestyle{3pt}{1.1}
\begin{tabular}{l|l|l|c|c|cc}
          method & backbone & decoder & T  & AP  & AR\textsubscript{1}  & AR\textsubscript{10}     \\
\shline
        \multirow{4}{*}{MaskProp~\cite{bertasius2020classifying}}
                & ResNet-50 + FPN~\cite{lin2017feature} + HTC~\cite{chen2019hybrid} & \multirow{4}{*}{\parbox{3cm}{Mask R-CNN~\cite{he2017mask} style\\ multi-heads \\ w/ mask refinement \\postprocessing}}  & 13  & {40.0}    & -         & -   \\
          
                & ResNet-101 + FPN~\cite{lin2017feature} + HTC~\cite{chen2019hybrid}  & & 13  & {42.5}    & -         & -   \\
        
                & ResNeXt-101~\cite{xie2017aggregated} + FPN~\cite{lin2017feature} + HTC~\cite{chen2019hybrid} & & 13  & {44.3}    & -         & -   \\
        
        &{ResNeXt-101~\cite{xie2017aggregated} + FPN~\cite{lin2017feature} + HTC~\cite{chen2019hybrid} + deform.STSN~\cite{dai2017deformable, bertasius2018object}}
        & &{13}  &{{46.6}}  &{-}  &{-}  \\ \hline

        \multirow{3}{*}{Seq Mask R-CNN~\cite{lin2021video}}
                & ResNet-50 + FPN~\cite{lin2017feature} & \multirow{3}{*}{\parbox{3cm}{Mask R-CNN~\cite{he2017mask} style\\ multi-heads \\w/ many proposals}}  & 36  & {40.4}    & 41.1     & 49.7   \\
          
                & ResNet-101 + FPN~\cite{lin2017feature}  & & 36  & {43.8}    & 46.3    & 52.6   \\
        
                & ResNeXt-101~\cite{xie2017aggregated} + FPN~\cite{lin2017feature} & & 36  & \textbf{47.6}    & 46.3    & 56.0   \\
        \hline
        
        \multirow{2}{*}{VisTR~\cite{wang2021end}}       
                &ResNet-50   & \multirow{2}{*}{\parbox{3.3cm}{DETR~\cite{carion2020end} style transformer}} & 36 & 36.2      & 37.2      & 42.4       \\
              
                & ResNet-101   & & 36 & 40.1      & 38.3      & 44.9       \\ \hline

        \multirow{3}{*}{IFC~\cite{hwang2021video}} 
                &ResNet-50 + FPN~\cite{lin2017feature}  & \multirow{3}{*}{\parbox{3.3cm}{DETR~\cite{carion2020end} style transformer}}  & 5  & 41.0      & 43.5      & 52.7       \\
        
                & ResNet-50 + FPN~\cite{lin2017feature} &  & 36 & 42.8      & 43.8      & 51.2       \\
        
                & ResNet-101 + FPN~\cite{lin2017feature} &  & 36 & 44.6      & 44.0      & 52.1       \\

\hline
        \multirow{2}{*}{STEm-Seg~\cite{athar2020stem}}  & ResNet-50 + FPN~\cite{lin2017feature}  
        & \multirow{2}{*}{\parbox{3.3cm}{3D Conv-based TSE~\cite{athar2020stem} \\ w/ multi-heads}} & 8  & 30.6      & 31.6      & 37.1        \\
        & ResNet-101 + FPN~\cite{lin2017feature}  & & 8  & 34.6      & 34.4      & 41.6        \\\hline\hline
        TF-DL-B1 (per-pixel)     
                & ResNet-50 + axial-attention~\cite{wang2020axial}$\dagger$  & \multirow{4}{*}{\parbox{3.3cm}{tube-transformer}}   & 5  & 36.4      & 40.8      & 49.5        \\
        {\color[HTML]{FFFFFF}TF-DL-B1} (per-mask)      
                & ResNet-50 + axial-attention~\cite{wang2020axial}$\dagger$ &  & 5  
                & 38.8      & 44.0      & 51.4        \\
        TF-DL-B4 (per-pixel)      
                & ResNet-50-n4 + axial-attention~\cite{wang2020axial}$\dagger$  & & 5  & {45.4} & {48.3} & {56.9}    \\
        {\color[HTML]{FFFFFF}TF-DL-B4} (per-mask)      
                & ResNet-50-n4 + axial-attention~\cite{wang2020axial}$\dagger$ &  & 5 
                & {47.5} & \textbf{50.2} & \textbf{59.0}    \\
\end{tabular}
\vspace{-.4em}
\caption{\textbf{[VIS]} YouTube-VIS-2019 {\it val} set results. $\dagger$: Axial attention blocks~\cite{wang2020axial} are used in the last two stages. ResNet-50-n4 scales the number of layers in stage-4 by 4 times (\ie, 24 blocks in total), resulting in a backbone with 104 layers.
}
\label{tab:sup_yt_vis_19}
\end{table*}
\section{Visualization}
\label{sec:visualization}

\begin{figure*}[!t]
\centering
\includegraphics[width=\linewidth]{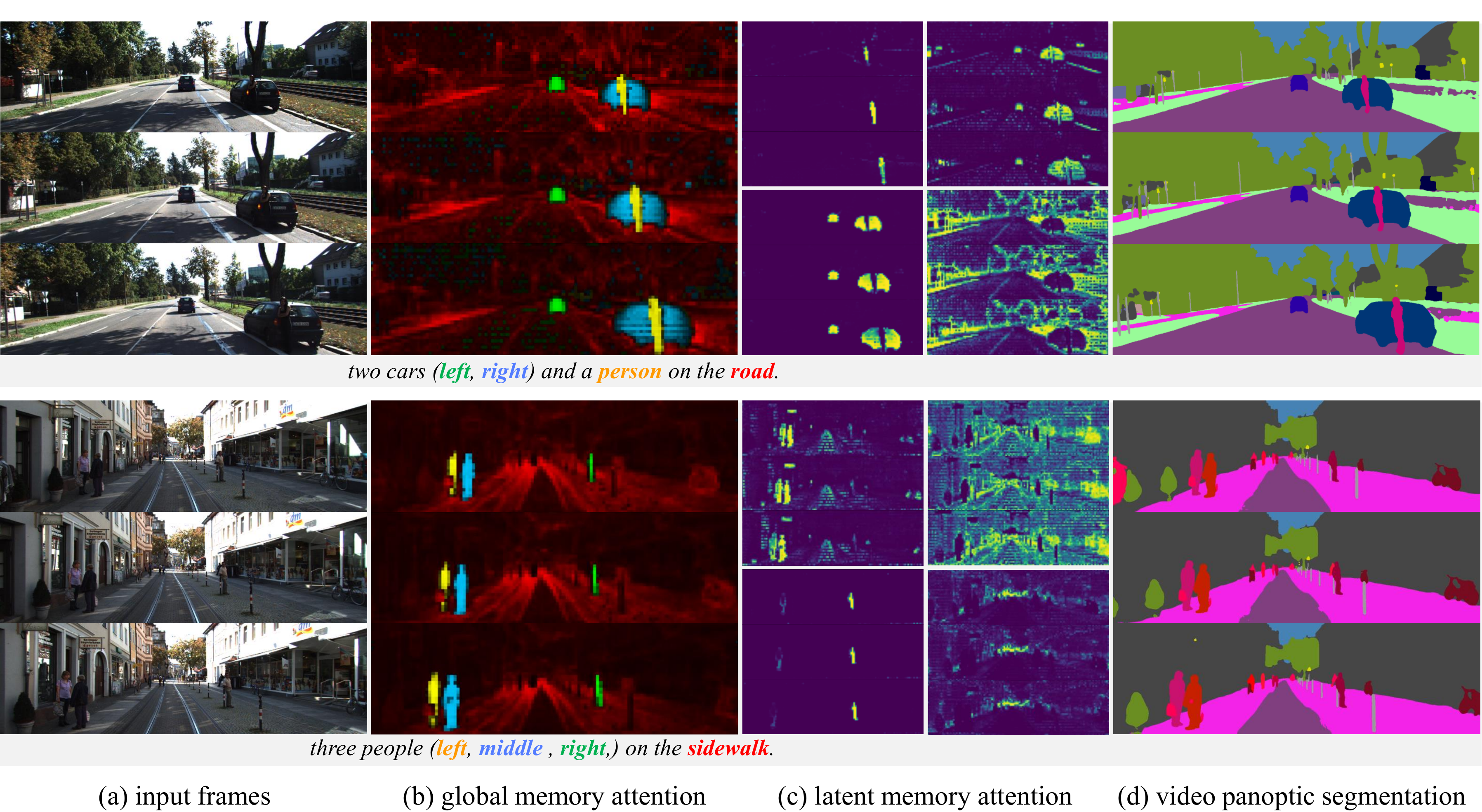}
\caption{\small \textbf{[VPS] Visualization on KITTI-STEP sequence}. From left to right: input frames ($T$=3), global memory attention, latent memory attention, and video {\it panoptic} segmentation results. The global memory attention is selected for predicted tube regions of interest, and the latent memory attention is selected for 4 (out of $L$=16)  latent memory.
}
\label{fig:visual_vps}
\end{figure*}

\begin{figure*}[t]
\centering
\includegraphics[width=\linewidth]{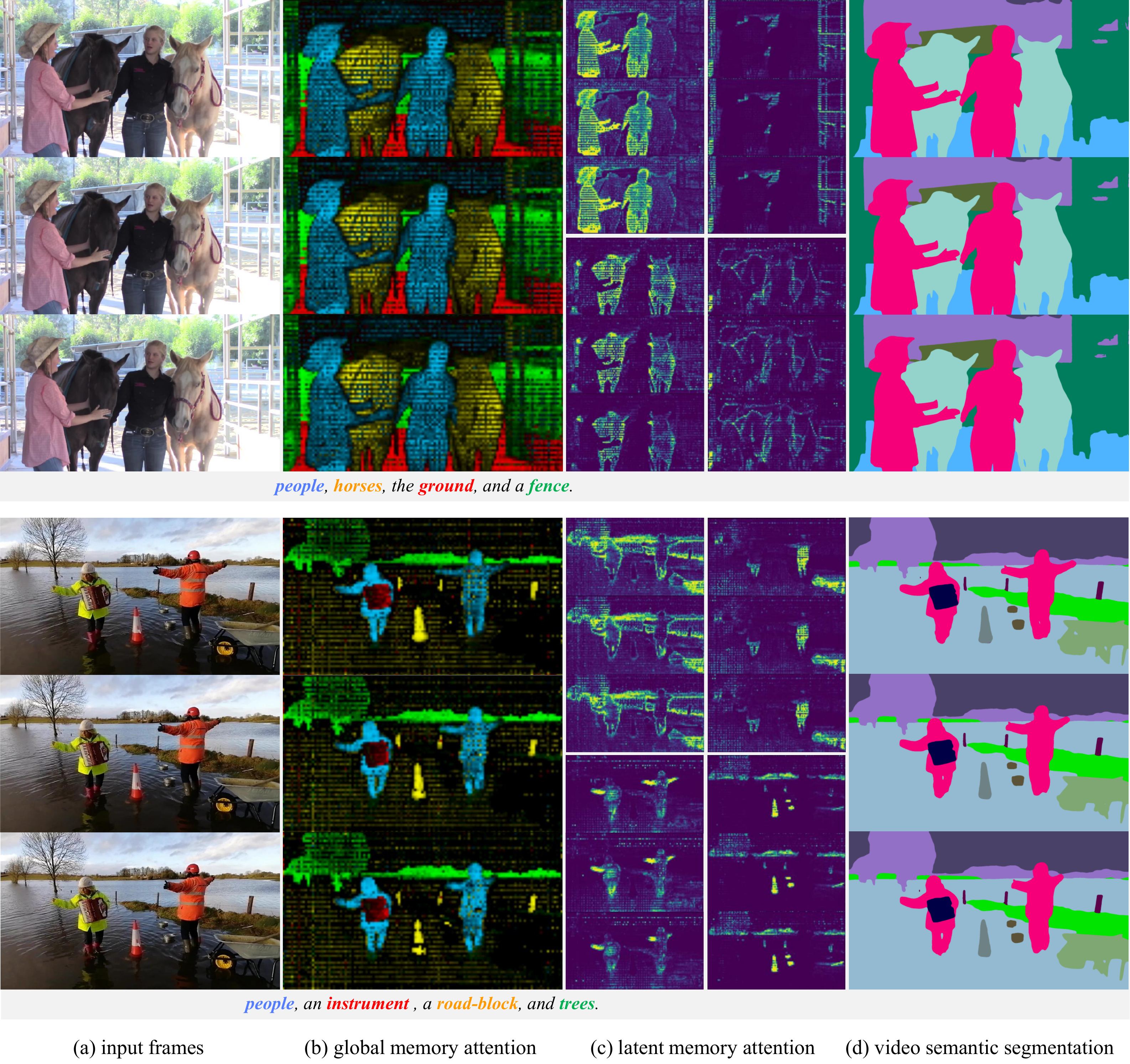}
\caption{\small \textbf{[VSS] Visualization on VSPW sequence}. From left to right: input frames ($T$=3), global memory attention, latent memory attention, and video {\it semantic} segmentation results. The global memory attention is selected for predicted tube regions of interest, and the latent memory attention is selected for 4 (out of $L$=16)  latent memory.
}
\label{fig:visual_vss}
\end{figure*}

\begin{figure*}[t]
\centering
\includegraphics[width=\linewidth]{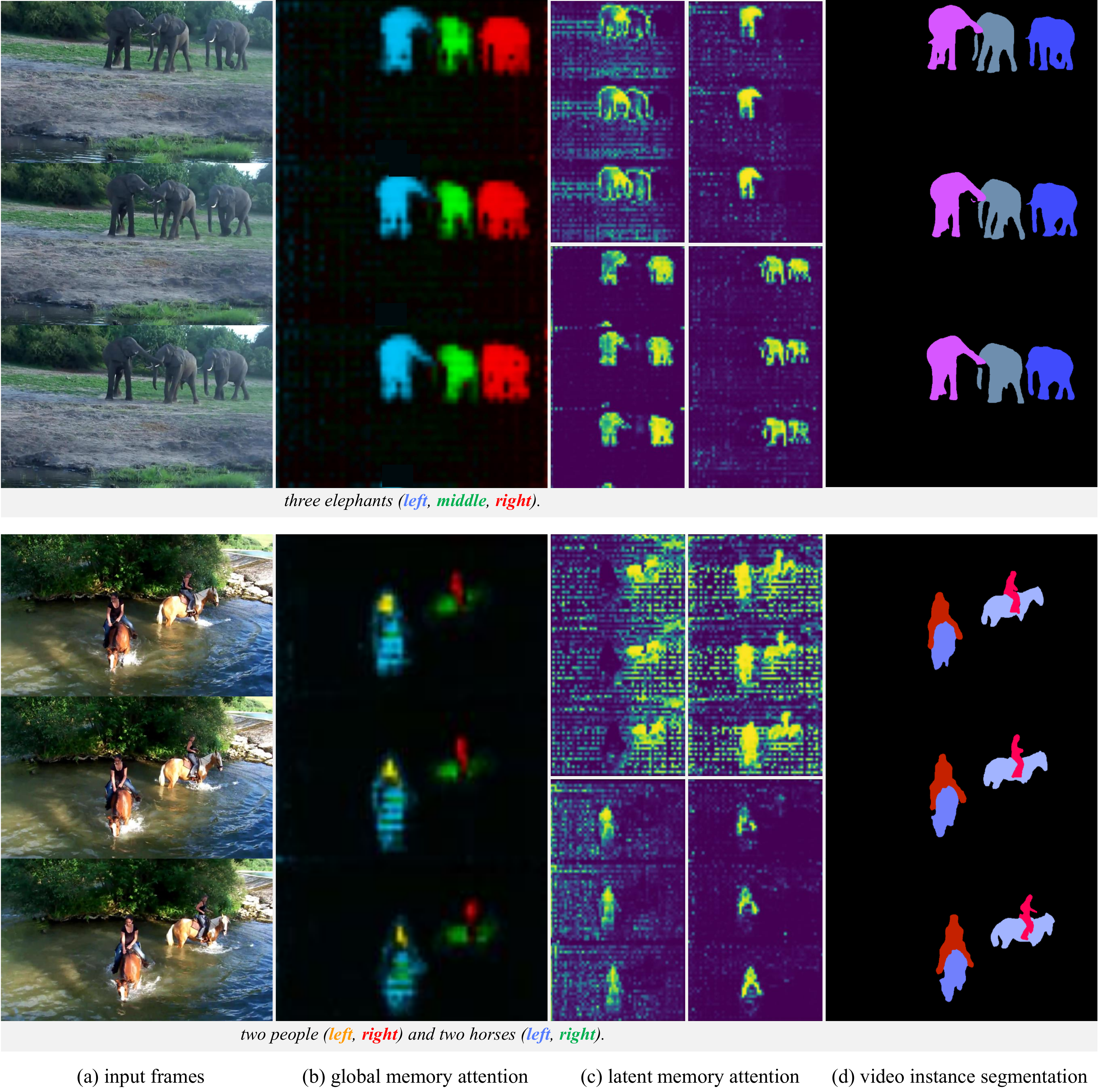}
\caption{\small \textbf{[VIS] Visualization on Youtube-VIS 2019 sequence}. From left to right: input frames ($T$=3), global memory attention, latent memory attention, and video {\it instance} segmentation results. The global memory attention is selected for predicted tube regions of interest, and the latent memory attention is selected for 4 (out of $L$=16)  latent memory.
}
\label{fig:visual_vis}
\end{figure*}

\begin{figure*}[t]
\centering
\includegraphics[width=\linewidth]{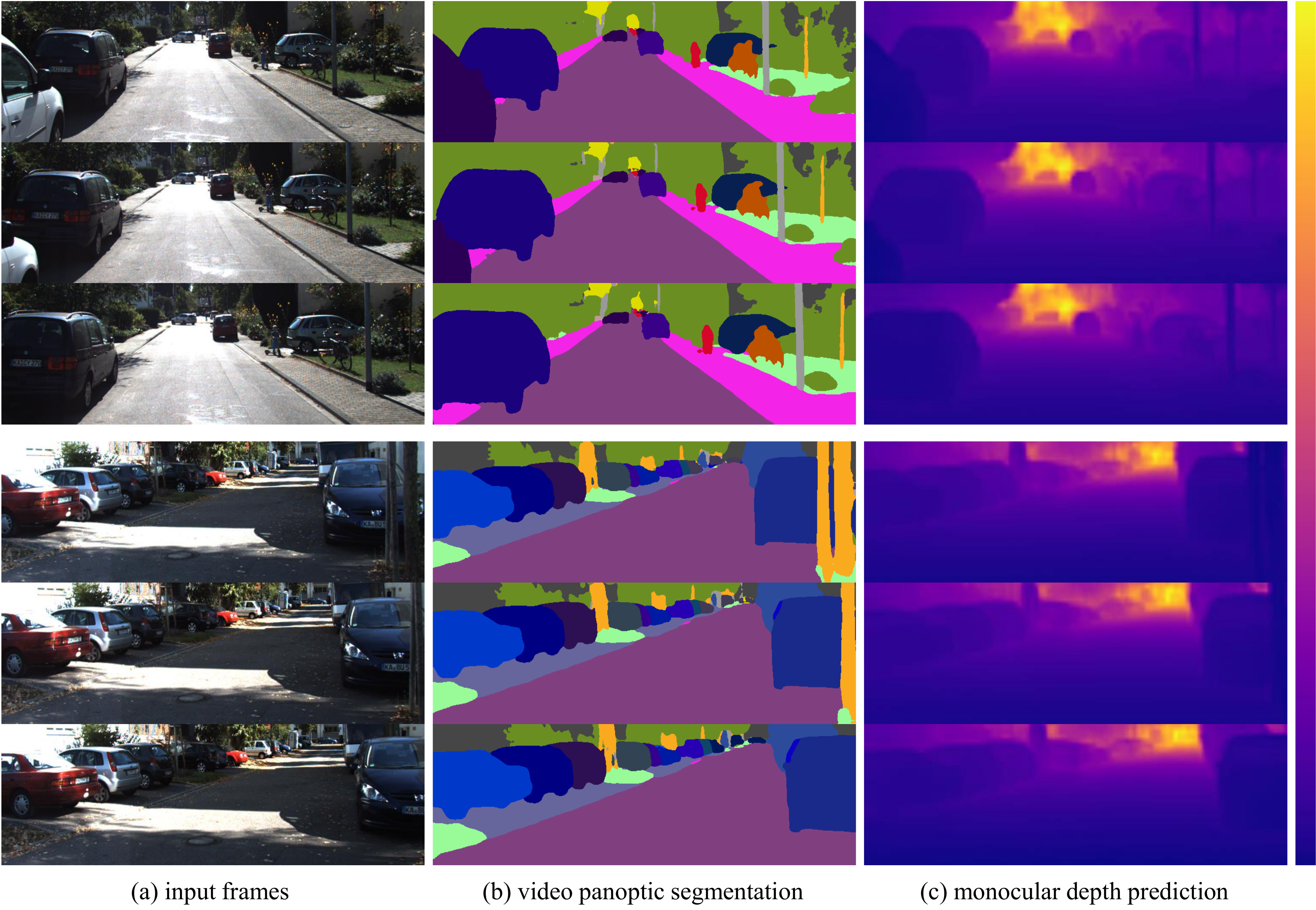}
\caption{\small \textbf{[DVPS] Visualization on SemKITTI-DVPS sequence}. From left to right: input frames ($T$=3), video \textit{panoptic} segmentation, and monocular depth prediction results. As the attentions are very similar to those in KITTI-STEP (\figref{fig:visual_vps}), here we focus on the depth visualization.
}
\label{fig:visual_dvps}
\end{figure*}

In \figref{fig:visual_vps}, \ref{fig:visual_vss}, and \ref{fig:visual_vis}, we visualize how the proposed hierarchical dual-path transformer performs attention for video panoptic/semantic/instance segmentation tasks (VPS, VSS, and VIS, respectively). We use input clips of three consecutive frames for visualization. For each sample, we select several output tubes of interest from the TubeFormer-DeepLab prediction. In column-{\color[HTML]{FF0000}b}, we probe the attention weights between the selected tube-specific global memory embeddings and all the pixels. Across all three tasks, we observe the global memory attention is spatio-temporally clustered for individual tube regions, while respecting different requirements among the tasks. That is, one global memory answers for each semantic category in VSS, but for each instance identity in VIS, while both cases appear in VPS task. 

In column-{\color[HTML]{FF0000}c}, we select four latent memory indices and visualize their attention maps. Commonly for all tasks, some latent memory learns to spatially specialize on certain areas (left \textit{vs} right side of the scene) or attends to the tube boundaries. Interestingly, we find that some latent memory focuses on relatively far-away region (\figref{fig:visual_vps}{\color[HTML]{FF0000}c}-bottom right), which often requires more attention. Sometimes, it has more interests to the moving object parts or small objects (\eg, \textit{moving arms} and \textit{a road-block cone} in \figref{fig:visual_vss}{\color[HTML]{FF0000}c}-bottom left and bottom right, respectively).

The task-specific behavior of the latent memory can be also compared between \figref{fig:visual_vss}{\color[HTML]{FF0000}c} and \figref{fig:visual_vis}{\color[HTML]{FF0000}c}. The latent memory in VSS does not distinguish instances of a same semantic class. In contrast, the attention is instance-specific in VIS. As shown in \figref{fig:visual_vis}{\color[HTML]{FF0000}c}-top left, the \textit{occluded noses of two elephants} are highlighted, which is expected to help the instance discrimination. Also, different latent memory attends to a single, or different multiples of the instances. 

Additionally, \figref{fig:visual_dvps} visualizes our depth-aware video panoptic segmentation results on SemKITTI-DVPS dataset, where TubeFormer-DeepLab is able to generate temporally consistent panoptic segmentation and monocular depth estimation results.

\section{Discussion}
\label{sec:discussion}
We notice that recently there is some hype in the literature regarding the development of \textit{universal} or \textit{unified} segmentation models for semantic, instance, and panoptic segmentation.
We would like to emphasize that the goal of panoptic segmentation is to unify semantic and instance segmentation, and thus a well-designed panoptic segmentation model should naturally demonstrate a fair performance on semantic segmentation and instance segmentation as well.
For example, Panoptic-DeepLab~\cite{cheng2019panoptic} and its Naive-Student version~\cite{chen2020naive} already demonstrate that a modern panoptic segmentation model could simultaneously achieve state-of-the-art performance on semantic, instance, and panoptic segmentation.
Our work follows the same direction by working on the video segmentation tasks.
\section{Limitations}

% We observe two main limitations in the proposed TubeFormer-DeepLab.

% \paragraph{Only short-term tracking.}\quad
Currently, the proposed TubeFormer-DeepLab performs clip-level video segmentation with the clip value $T=2$ (for VPS and VSS) or $T=5$ (for VIS). Our model thus performs short-term tracking and may miss objects that have track lengths larger than the used clip value. This limitation is also reflected in the AQ (association quality) reported in Tab. 1 of the main paper (\ie, KITTI-STEP {\it test} set results). We leave the question about how to efficiently incorporate long-term tracking to TubeFormer-DeepLab for feature work.

% \paragraph{Only one prediction per pixel.}\quad
% Our model TubeFormer-DeepLab only generates one prediction for each pixel in the final output. This facilitates video segmentation tasks that only require one prediction per pixel (\eg, video panoptic segmentation and video semantic segmentation) without the need to design extra modules to resolve the overlapping predictions. However, our model performs worse on some video segmentation tasks that favor proposal-based methods (\eg, Youtbue-VIS dataset). It is therefore interesting to adapt the TubeFormer-DeepLab for tube proposals, \eg via using a feature pyramid network~\cite{lin2017feature}.

In any case, our proposed TubeFormer-DeepLab presents the first attempt to tackle multiple video segmentation tasks from a unified approach. We hope our simple and effective model could serve as a solid baseline for future research.
\section{Conclusion}
\label{sec:conclusion}

We introduced TubeFormer-DeepLab, a novel architecture based on mask transformers for video segmentation. Video segmentation tasks, particularly video semantic/instance/panoptic segmentation, have been tackled by fundamentally divergent models. We proposed a new paradigm that formulates video segmentation tasks as the problem of partitioning video tubes with different predicted labels. TubeFormer-DeepLab, directly predicting class-labeled tubes, provides a general solution to multiple video segmentation tasks.
We hope our approach will inspire future research in the unification of video segmentation tasks.

% \clearpage
%%%%%%%%% REFERENCES
{\small
\bibliographystyle{ieee_fullname}
\bibliography{arxiv}
}

\end{document}